\documentclass[hyperref,UTF8,11px,a4paper]{article} 

\usepackage[margin=1in]{geometry} 

\usepackage{graphicx} 
\graphicspath{{image/}} 
\usepackage{epstopdf}
\usepackage{abstract} 
\usepackage{amsmath}
\usepackage{amsthm}
\usepackage{amsfonts}
\usepackage{amssymb}
\usepackage{url}
\usepackage{hyperref}
\hypersetup{
citecolor=blue,
pdfstartview=FitH,bookmarksnumbered=true,bookmarksopen=true,} 
\usepackage{booktabs} 
\usepackage[authoryear,round]{natbib}
\usepackage{array} 
\usepackage{paralist} 
\usepackage{verbatim} 
\usepackage{subfigure} 
\usepackage{cases} 
\usepackage{multirow} 
\usepackage{lastpage} 
\usepackage{algorithm}
\usepackage{algorithmic}
\usepackage{comment} 
\usepackage{listings}
\usepackage{fancyhdr} 
\usepackage{sectsty}
\allsectionsfont{\sffamily\mdseries\upshape} 

\usepackage[notlot,notlof]{tocbibind} 
\usepackage[titles,subfigure]{tocloft} 

\title{Incorporating Transformer and LSTM to Kalman Filter with EM algorithm for state estimation}
\author{Zhuangwei Shi${}^{1,*}$\\
    ${}^1$College of Artificial Intelligence, Nankai University, Tongyan Road, Tianjin, 300350, China\\
    $^\ast$Correspondence: \url{zwshi@mail.nankai.edu.cn}
}

\begin{document}
\maketitle
\thispagestyle{fancy} 

\section*{Abstract}

Kalman Filter requires the true parameters of the model and solves optimal state estimation recursively. Expectation Maximization (EM) algorithm is applicable for estimating the parameters of the model that are not available before Kalman filtering, which is EM-KF algorithm. To improve the preciseness of EM-KF algorithm, the author presents a state estimation method by combining the Long-Short Term Memory network (LSTM), Transformer and EM-KF algorithm in the framework of Encoder-Decoder in Sequence to Sequence (seq2seq). Simulation on a linear mobile robot model demonstrates that the new method is more accurate. Source code of this paper is available at \url{https://github.com/zshicode/Deep-Learning-Based-State-Estimation}.

\section*{Keywords}

Kalman Filter, State Estimation, Expectation Maximization (EM), Long
Short-Term Memory (LSTM), Transformer

\section{Introduction}

State estimation is an important problem in control theory and machine learning, and is widely applied for robotics, computer vision, time-series forecasting and so on. As the state and observation of a system are usually interfered by stochastic noise, state estimation aims to estimate the true state via observation, and minimize the error between estimation and true state. \citet{kalman1960} proposed Kalman Filter (KF) to recursively estimate state via observation in stochastic linear systems.
\begin{equation}
    \begin{split}
        x_k&=Ax_{k-1}+w_{k},\\
        y_k&=Cx_{k}+v_{k},\\
        w_{k}&\sim N(0,Q),v_{k}\sim N(0,R),\\
        x_{0}&\sim N(m_{0},P_{0}),\, k=1,2,...,N,
    \end{split}\label{eq:ssm} 
\end{equation}
Basic Kalman Filter was designed for linear systems. Also, there are some nonlinear extensions of Kalman Filter, such as Extended Kalman Filter (EKF), Unscented Kalman Filter (UKF, also known as sigma-point Kalman Filter) and Particle Filter (PF) \citep{barfoot2017}.

Although Kalman Filter and its extensions are widely used for state estimation, there are two disadvantages of KF that cannot be ignored.
\begin{enumerate}
    \item KF requires model parameters. Although $A,C$ are easy to obtain through system modeling, $Q,R,m_0,P_0$ depend on estimation by designers' experience.
    \item Eq. \eqref{eq:ssm} clarifies that KF assumpts Markov property of states, and conditional independence of observations, yet de facto systems do not follow these two assumptions usually. 
\end{enumerate}

To address the first issue, Expectation Maximization (EM) algorithm \citep{em1977} was used for parameter estimation before filtering. \citet{emkf1982} first proposed a method that adopted EM algorithm to estimate parameters for Kalman Filter. Based on this work, \citet{emkf1996} proposed a method integrating Kalman Filter for state estimation and EM algorithm for parameter estimation in linear dynamic systems, and it is so called EM-KF algorithm. The EM-KF algorithm is also extended to nonlinear systems, such as EM-EKF \citep{emekf2012}, EM-UKF \citep{merwe2003} and EM-PF \citep{empf2008,empf2015}.

To address the second issue, deep learning based approaches were proposed to state estimation. State estimation can also be seen as estimation on sequences. Recurrent Neural Network (RNN) was adopted for analyzing sequential data \citep{Elman1990rnn}, and Long Short-Term Memory (LSTM) \citep{lstm1997} model is the most commonly used RNN. Recently, there are many works Incorporating LSTM to state estimation \citep{Wan2018Multi,Yeo2018Deep,lstm2018velocity}.

LSTM introduced gate mechanism in RNN, which can be seen as simulation for human memory, that human can remember useful information and forget useless information. Attention Mechanism \citep{attention2015} can be seen as simulation for human attention, that human can pay attention to useful information and ignore useless information. Transformer \citep{transformer2017} is a self-attention based sequence-to-sequence (seq2seq) \citep{seq2seq2014} model to encode and decode sequential data. Experiments on natural language processing demonstrates that Transformer can solve long-term dependency problem in LSTM, hence, Transformer can better model long sequences. LSTM can capture particular long-distance correspondence that fits the sturcture of LSTM itself, while Transformer can capture much longer correspondence. Therefore, Transformer is more flexible and robust.

In this paper, we proposed an encoder-decoder framework in seq2seq for state estimation, that state estimation is equivalent to encode and decode observation. Previous works incorporating LSTM to KF, are adopting LSTM encoder and KF decoder. Here, we proposed LSTM-KF adopting LSTM encoder and EM-KF decoder. Before EM-KF decoder, we replaced LSTM encoder by Transformer encoder, and we called this Transformer-KF. Then, we integrated Transformer and LSTM, and it was so called TL-KF.Experiments demonstrate that integrating Transformer and LSTM to encode observation before filtering, makes it easier for EM algorithm to estimate parameters. Source code of this paper is available at \url{https://github.com/zshicode/Deep-Learning-Based-State-Estimation}.

\section{Material and Methods}

\subsection{Kalman Filter and Kalman Smoother}

Let $y_{1:k}$ denotes the observation of a system at $1\sim k$, when $y_{1:k}$ and parameters of Eq. \eqref{eq:ssm}, $A,C,Q,R,m_0,P_0$, have already been known, Kalman Filter recursively and forwardly estimates $x_k$ from $x_{k-1}$.
\begin{equation}
    p(x_k|y_{1:k})=N(x_k|m_{k|k},P_{k|k}).
\end{equation}
Here, $N(x|m,P)$ denotes that $x$ follows a Gaussian distribution, whose mean value is $m$ and covarience matrix is $P$. Kalman Filter is an online state estimation algorithm, when $y_{1:N}$ (i.e. observation of the whole process), along with model parameters, have already been known, state estimation can be obtained via offline way, which is more precise. This is Kalman Smoother (KS) \citep{rts1965}, also known as Rauch-Tung-Striebel Smoother or RTS Smoother. Based on the computation of Kalman Filter, Kalman Smoother estimates the state recursively and forwardly, for each $k=N-1,N-2,...,0$.
\begin{equation}
    p(\hat x_k|y_{1:N})=N(\hat x_k|m_{k|N},P_{k|N}),\label{eq:ks}
\end{equation}
The state estimation obtained through Kalman Filter or Kalman Smoother, is least square estimation (LSE) from control and optimization perspective, along with maximum likelihood estimation (MLE) and maximum a posteriori estimation (MAP) from statistics and probability perspective \citep{Robert2012MLAPP,barfoot2017}.

Kalman Filter recursively obtains the optimal estimation of $x_k$ through the following steps.
\begin{enumerate}
    \item Initialization. $m_{0|0}=m_0,P_{0|0}=P_0.$
    \item For $k=1,2,...,N$, from the estimation at moment $k-1$, when the state is with mean value $m_{k-1|k-1}$ and variance $P_{k-1|k-1}$, to compute the one-step prediction from $k-1$ to $k$.
    \begin{equation}
        p(x_k|y_{1:k-1})=N(x_k|m_{k|k-1},P_{k|k-1}),
    \end{equation}
    where mean value and variance are
    \begin{equation}
        m_{k|k-1}=Am_{k-1|k-1},
    \end{equation}
    \begin{equation}
        P_{k|k-1}=AP_{k-1|k-1}A^T+Q,
    \end{equation}
    respectively.
    \item The optimal state estimation of $m_{k|k}$, is the linear combination of $m_{k|k-1}$, along with residual $r_k=y_k-Cm_{k|k-1}$ (also known as innovation).
    \begin{equation}
        m_{k|k}=m_{k|k-1}+H_k(y_k-Cm_{k|k-1}),
    \end{equation}
    where $H_k$ is Kalman Filter gain.
    \item When deriving least square estimation, $H_k$ can be obtained through solving equation $\frac{\partial \mathrm{tr}P_{k|k}}{\partial H_k}=0$, then $P_{k|k}$ can be obtained as well. When deriving maximum a posteriori estimation and maximum likelihood estimation, $H_k$ and $P_{k|k}$ can be obtained through Bayes theorem. (See Section 3.3.2 \emph{Kalman Filter via MAP}, Section 3.3.3 \emph{Kalman Filter via Bayesian Inference} and Section 3.3.4 \emph{Kalman Filter via Gain Optimization} of \citep{barfoot2017}.)
    \begin{equation}
        H_k=P_{k|k-1}C^T(CP_{k|k-1}C^T+R)^{-1},
    \end{equation}
    \begin{equation}
        P_{k|k}=(I-H_kC)P_{k|k-1}.
    \end{equation}
    \item Finally, $m_{k|k}$ is the optimal filtering estimation of $x_k$.
    \begin{equation}
        p(x_k|y_{1:k})=N(x_k|m_{k|k},P_{k|k}).
    \end{equation}
\end{enumerate}

Similarly, in Kalman Smoother
\begin{equation}
    p(\hat x_k|y_{1:N})=N(\hat x_k|m_{k|N},P_{k|N}),\label{eq:ks}
\end{equation}
where
\begin{equation}
    m_{k|N}=m_{k|k}+J_k (m_{k+1|N}-m_{k+1|k}),
\end{equation}
\begin{equation}
    P_{k|N}=P_{k|k}+J_k (P_{k+1|N}-P_{k+1|k}) J_k^T.
\end{equation}
Kalman smoother gain (See Section 3.2.3 \emph{Rauch-Tung-Striebel Smoother} of \citep{barfoot2017})
\begin{equation}
    J_k=P_{k|k}A^T P^{-1}_{k+1|k}.
\end{equation}
In a word, for $k=N-1,N-2,...,0$, Kalman Smoother iterates optimal state estimation at the opposite direction of Kalman Filter.

\subsection{EM-KF algorithm}

Expectation Maximization (EM) algorithm iteratively computes the maximum likelihood estimation (MLE) of parameters when the probability distribution function (PDF) is with latent variables. According to Jenson's inequality,
\begin{equation}
    \begin{split}
        \log p\left(\mathbf{y}_{1 : N} | \boldsymbol{\theta}\right) &=\log \left(\int q\left(\mathbf{x}_{0 : N}\right) \frac{p\left(\mathbf{x}_{0 : N}, \mathbf{y}_{1 : N} | \boldsymbol{\theta}\right)}{q\left(\mathbf{x}_{0 : N}\right)} \mathrm{d} \mathbf{x}_{0 : N}\right)\\&\geq \int q\left(\mathbf{x}_{0 : N}\right) \log \frac{ p\left(\mathbf{x}_{0 : N}, \mathbf{y}_{1 : N} | \boldsymbol{\theta}\right)}{q\left(\mathbf{x}_{0 : N}\right)} \mathrm{d} \mathbf{x}_{0 : N}.
    \end{split}\label{eq:jenson}
\end{equation}
Let $\theta^{(n)}$ denotes the parameters at $n$-th iteration, the PDF of latent variables
\begin{equation} 
q\left(\mathbf{x}_{0 : N}\right) :=p\left(\mathbf{x}_{0 : N} | \mathbf{y}_{1 : N}, \boldsymbol{\theta}^{(n)}\right),\label{eq:q}
\end{equation}
Substitute Eq. \eqref{eq:q} into Eq. \eqref{eq:jenson},
\begin{equation}
    \begin{split}
        &\int p\left(\mathbf{x}_{0 : N} | \mathbf{y}_{1 : N}, \boldsymbol{\theta}^{(n)}\right) \log \frac{p\left(\mathbf{x}_{0 : N}, \mathbf{y}_{1 : N} | \boldsymbol{\theta}\right)}{p\left(\mathbf{x}_{0 : N} | \mathbf{y}_{1 : N}, \boldsymbol{\theta}^{(n)}\right)} \mathrm{d} \mathbf{x}_{0 : N}\\=&\int p\left(\mathbf{x}_{0 : N} | \mathbf{y}_{1 : N}, \boldsymbol{\theta}^{(n)}\right) \log p\left(\mathbf{x}_{0 : N}, \mathbf{y}_{1 : N} | \boldsymbol{\theta}\right) \mathrm{d} \mathbf{x}_{0 : N}\\&-\int p\left(\mathbf{x}_{0 : N} | \mathbf{y}_{1 : N}, \boldsymbol{\theta}^{(n)}\right) \log p\left(\mathbf{x}_{0 : N} | \mathbf{y}_{1 : N}, \boldsymbol{\theta}^{(n)}\right) \mathrm{d} \mathbf{x}_{0 : N}.
    \end{split}\label{eq:maxlikelihood}
\end{equation}
Note that the latter is independent to $\theta$, it can be omitted \citep{shi2021vgaelda}. For state estimation, maximizing the loglikelihood of the PDF of observation $\log p\left(\mathbf{y}_{1 : N} | \boldsymbol{\theta}\right)$, is equivalent to maximize
\begin{equation} 
    \mathcal{Q}\left(\boldsymbol{\theta}, \boldsymbol{\theta}^{(n)}\right)=\mathbb{E}\left[\log p\left(\mathbf{x}_{0 : N}, \mathbf{y}_{1 : N} | \boldsymbol{\theta}\right) | \mathbf{y}_{1 : N}, \boldsymbol{\theta}^{(n)}\right].\label{eq:estep}
\end{equation}
EM algorithm is begin with the initial estimation $\theta^{(0)}$, then iteratively executes E-step and M-step as Algorithm \ref{alg:em}.

\begin{algorithm}[H]
    \caption{EM algorithm}\label{alg:em}
    \begin{algorithmic}[1]
        \REQUIRE Initial estimation of parameters $\boldsymbol{\theta}^{(0)}$, error $\epsilon$, maximum iteration number $n_m$
        \ENSURE $\boldsymbol{\theta}^{*}$
        \REPEAT
        \STATE E-step: compute $\mathcal{Q}\left(\boldsymbol{\theta}, \boldsymbol{\theta}^{(n)}\right)$
        \STATE M-step: $\boldsymbol{\theta}^{(n+1)} \leftarrow \arg \max _{\boldsymbol{\theta}} \mathcal{Q}\left(\boldsymbol{\theta}, \boldsymbol{\theta}^{(n)}\right)$
        \UNTIL $|\boldsymbol{\theta}^{(n+1)}-\boldsymbol{\theta}^{(n)}|<\epsilon$, or iteration number is up to $n_m$
        \RETURN $\boldsymbol{\theta}^{*}\leftarrow \boldsymbol{\theta}^{(n+1)}$
    \end{algorithmic}
\end{algorithm}

For linear system \eqref{eq:ssm}, the probability distribution is
\begin{subequations}
    \begin{gather}
        p\left(\mathbf{x}_{k} | \mathbf{x}_{k-1}\right)= N(\mathbf{x}_{k}|A \mathbf{x}_{k-1},Q),\\p\left(\mathbf{y}_{k} | \mathbf{x}_{k}\right)=N(\mathbf{y}_{k}|C \mathbf{x}_{k},R),\\p\left(\mathrm{x}_{0}\right)=N({m}_{0},{P}_{0}).
    \end{gather}
\end{subequations}
Here,
\begin{equation} 
p\left(\mathbf{x}_{k} | \mathbf{x}_{k-1}\right)=\exp \left\{-\frac{1}{2}\left[\mathbf{x}_{k}-A \mathbf{x}_{k-1}\right]^{T} Q^{-1}\left[\mathbf{x}_{k}-A \mathbf{x}_{k-1}\right]\right\}(2 \pi)^{-u / 2}|Q|^{-1 / 2}
\end{equation}
\begin{equation} 
p\left(\mathbf{y}_{k} | \mathbf{x}_{k}\right)=\exp \left\{-\frac{1}{2}\left[\mathbf{y}_{k}-C \mathbf{x}_{k}\right]^{T} R^{-1}\left[\mathbf{y}_{k}-C \mathbf{x}_{k}\right]\right\}(2 \pi)^{-v / 2}|R|^{-1 / 2}
\end{equation}
\begin{equation} 
p\left(\mathrm{x}_{0}\right)=\exp \left\{-\frac{1}{2}\left[\mathrm{x}_{0}-{m}_{0}\right]^{\prime} {P}_{0}^{-1}\left[\mathrm{x}_{0}-{m}_{0}\right]\right\}(2 \pi)^{-u / 2}\left|{P}_{0}\right|^{-1 / 2}
\end{equation}
where $u,v$ denote the dimension of state vector and observation vector, respectively. With the assumption of Markov property of states, and conditional independence of observations,
\begin{equation} 
p(\mathbf{x}_{0:N},\mathbf{y}_{1:N})=p\left(\mathbf{x}_{0}\right) \prod_{k=1}^{N} p\left(\mathbf{x}_{k} | \mathbf{x}_{k-1}\right) \prod_{k=1}^{N} p\left(\mathbf{y}_{k} | \mathbf{x}_{k}\right).
\end{equation}
Hence,
\begin{equation} 
\begin{aligned} \ln p\left(\mathbf{x}_{0 : N}, \mathbf{y}_{1 : N}\right)=&-\sum_{k=1}^{N}\left(\frac{1}{2}\left[\mathrm{y}_{k}-C \mathrm{x}_{k}\right]^{T} R^{-1}\left[\mathrm{y}_{k}-C \mathrm{x}_{k}\right]\right)-\frac{N}{2} \ln |2\pi R| \\ &-\sum_{k=1}^{N}\left(\frac{1}{2}\left[\mathrm{x}_{k}-A \mathrm{x}_{k-1}\right]^{T} Q^{-1}\left[\mathrm{x}_{k}-A \mathrm{x}_{k-1}\right]\right)-\frac{N}{2} \ln |2\pi Q| \\ &-\frac{1}{2}\left[\mathrm{x}_{1}-{m}_{0}\right]^{T} {P}_{0}^{-1}\left[\mathrm{x}_{1}-{m}_{0}\right]-\frac{1}{2} \ln \left|2\pi {P}_{0}\right|,\end{aligned}
\end{equation}

Substitute Eq. \eqref{eq:ssm} into Eq. \eqref{eq:estep},
\begin{equation}
    \begin{split}\mathcal{Q}\left(\boldsymbol{\theta}, \boldsymbol{\theta}^{(n)}\right)=&-\frac{1}{2} \ln \left|2 \pi \mathbf{P}_{0}\right|-\frac{N}{2} \ln |2 \pi \mathbf{Q}|-\frac{N}{2} \ln |2 \pi \mathbf{R}| \\ &{-\frac{1}{2} \mathrm{tr}\left\{\mathbf{P}_{0}^{-1}\left[\mathbf{P}_{0 | N}+\left(\mathbf{m}_{0 | N}-\mathbf{m}_{0}\right)\left(\mathbf{m}_{0 | N}-\mathbf{m}_{0}\right)^{T}\right]\right\}} \\ &{-\frac{1}{2} \sum_{k=1}^{N} \mathrm{tr}\left\{\mathbf{Q}^{-1} \mathbb{E}\left[\left(\mathbf{x}_{k}-\mathbf{Ax}_{k-1}\right)\left(\mathbf{x}_{k}-\mathbf{Ax}_{k-1}\right)^{T} | \mathbf{y}_{1 : N}\right]\right\}}\\&-\frac{1}{2} \sum_{k=1}^{N} \mathrm{tr}\left\{\mathbf{R}^{-1} \mathbb{E}\left[\left(\mathbf{y}_{k}-\mathbf{C}\mathbf{x}_k\right)\left(\mathbf{y}_{k}-\mathbf{C}\mathbf{x}_k\right)^{T} | \mathbf{y}_{1 : N}\right]\right\}.\end{split}
\end{equation}
Then, solve equation
\begin{equation}
    \frac{\partial \mathcal{Q}\left(\boldsymbol{\theta}, \boldsymbol{\theta}^{(n)}\right)}{\partial \boldsymbol{\theta}^{(n)}}=0.
\end{equation}
For $A,C,m_0$, we have,
\begin{equation}
    A = \left( \sum_{k=1}^{N-1} \mathbb{E}[\hat x_k \hat x_{k-1}^T]\right)
            \left( \sum_{k=1}^{N-1} \mathbb{E}[\hat x_{k-1} \hat x_{k-1}^T] \right)^{-1},
\end{equation}
\begin{equation}
    C = \left( \sum_{k=0}^{N-1} y_k \mathbb{E}[\hat x_k]^T \right)
            \left( \sum_{k=0}^{N-1} \mathbb{E}[\hat x_k \hat x_k^T] \right)^{-1},
\end{equation}
\begin{equation}
    m_0=\mathbb{E}[\hat x_0].
\end{equation}
Then, compute $Q,R,P_0$ via $A,C,m_0$.
\begin{equation}
    \begin{split}
        Q =& \frac{1}{N-1} \sum_{k=0}^{N-2}
        (\mathbb{E}[\hat x_{k+1}] - A \mathbb{E}[\hat x_k])
            (\mathbb{E}[\hat x_{k+1}] - A \mathbb{E}[\hat x_k])^T\\
        &+ A \mathrm{Var}[\hat x_k] A^T +\mathrm{Var}[\hat x_{k+1}]\\
        &- \mathrm{Cov}(\hat x_{k+1}, \hat x_k) A^T - A \mathrm{Cov}(\hat x_k, \hat x_{k+1}),
    \end{split}
\end{equation}
\begin{equation}
    R= \frac{1}{N} \sum_{k=0}^{N-1}
                [z_k - C \mathbb{E}[\hat x_k]]
                    [z_k - C \mathbb{E}[\hat x_k]]^T
                + C \mathrm{Var}[\hat x_k] C^T,
\end{equation}
\begin{equation}
    P_0=\mathbb{E}[\hat x_0, \hat x_0^T] - m_0 \mathbb{E}[\hat x_0]^T
                - \mathbb{E}[\hat x_0] m_0^T + m_0 m_0^T.
\end{equation}
Here, $\hat x_k$ denotes the estimation of Kalman Smoother as Eq. \eqref{eq:ks}.

\subsection{Deep learning for sequential data}

\subsubsection{Long-short term memory model}

In basic feed-forward neural network (FFNN), output of current moment $o_t$ is only determined by input of current moment $i_t$, which suppress the ability of FFNN to  model time-series data. In recurrent neural network (RNN), a delay is used to save the latent state of latest moment $h_{t-1}$, then, latent state of current moment $h_t$ is determined by both $h_{t-1}$ and $i_t$. \citet{lstm1997} suggested that RNN may vanish the gradient as error propagates through time dimension, which leads to long-term dependency problem. Human can selectively remember information. Through gated activation function, LSTM (long short-term memory) model can selectively remember updated information and forget accumulated information.

\subsubsection{Seq2seq model}

Sequence-to-sequence model (seq2seq) \citep{seq2seq2014} is constructed through an encoder-decoder architecture, which enhances the ability of LSTM to learn hidden information through data with noise. In seq2seq, the encoder is an LSTM that encodes the inputs into the context (commonly the hidden state at last $h_N$), then decode the context in the decoder. In the decoder, output at previous moment is input at next moment.

\subsubsection{Attention mechanism}

Human usually pays attention to salient information. Attention mechanism is a technique in deep learning based on human cognitive system. For input $X=(x_1,x_2,...,x_N)$, give query vector $q$, depict the index of selected information by attention $z=1,2,...,N$, then the distribution of attention
\begin{equation} 
\alpha_{i} =p(z=i | X, \mathbf{q}) =\mathrm{softmax}\left(s\left(\mathbf{x}_{i}, \mathbf{q}\right)\right) =\frac{\exp \left(s\left(\mathbf{x}_{i}, \mathbf{q}\right)\right)}{\sum_{j=1}^{N} \exp \left(s\left(\mathbf{x}_{j}, \mathbf{q}\right)\right)}.
\end{equation}
Here
\begin{equation} 
s\left(\mathbf{x}_{i}, \mathbf{q}\right)=\frac{\mathbf{x}_{i}^{\mathrm{T}} \mathbf{q}}{\sqrt{d}}
\end{equation}
is attention score through scaled dot product, $d$ is the dimension of input. Suppose the input key-value pairs $(K, V)=\left[\left(\mathbf{k}_{1}, \mathbf{v}_{1}\right), \cdots,\left(\mathbf{k}_{N}, \mathbf{v}_{N}\right)\right]$, for given $q$, attention function
\begin{equation} 
\mathrm{att}((K, V), \mathbf{q}) =\sum_{i=1}^{N} \alpha_{i} \mathbf{v}_{i} =\sum_{i=1}^{N} \frac{\exp \left(s\left(\mathbf{k}_{i}, \mathbf{q}\right)\right)}{\sum_{j} \exp \left(s\left(\mathbf{k}_{j}, \mathbf{q}\right)\right)} \mathbf{v}_{i} .
\end{equation}

Multi-head mechanism is usually adopted through multi-query $Q=\left[\mathbf{q}_{1}, \cdots, \mathbf{q}_{M}\right]$ for attention function computation.
\begin{equation} 
\mathrm{att}((K, V), Q)=\mathrm{Concatenate}\left(\mathrm{att}\left((K, V), \mathbf{q}_{1}\right),\cdots,\mathrm{att}\left((K, V), \mathbf{q}_{M}\right)\right).
\end{equation}
This is so called multi-head attention (MHA).

Attention mechanism can be adopted to generate data-driven different weights. Here, $Q,K,V$ are all obtained through linear transform of $X$, and $W_Q,W_K,W_V$ can be adjusted dynamicly.
\begin{equation} 
Q =W_{Q} X,K =W_{K} X,V =W_{V} X.\label{eq:selfatt}
\end{equation}
This is so called self-attention. Similarly, output
\begin{equation} 
\mathbf{h}_{i} =\operatorname{att}\left((K, V), \mathbf{q}_{i}\right) =\sum_{j=1}^{N} \alpha_{i j} \mathbf{v}_{j} =\sum_{j=1}^{N} \operatorname{softmax}\left(s\left(\mathbf{k}_{j}, \mathbf{q}_{i}\right)\right) \mathbf{v}_{j} .
\end{equation}
Adopting scaled dot product score, the output
\begin{equation} 
H=V \operatorname{softmax}\left(\frac{K^{\mathrm{T}} Q}{\sqrt{d}}\right).\label{eq:selfatt:output}
\end{equation}

\subsubsection{Transformer}

Transformer solves long-term depedency via attention mechanism. Transformer implements encoder-decoder framework as seq2seq based on self-attention. Both encoder and decoder in Transformer consists of 6 blocks, and the output of last encoder block is the input of each decoder block. In Transformer, an encoder block consists of self-attention layer and FFNN, a decoder block consists of self-attention layer, encoder-decoder attention layer and FFNN. $Q,K,V$ are computed through Eq. \eqref{eq:selfatt} after self-attention layer, and $H$ is computed through Eq. \eqref{eq:selfatt:output}. This is the input of FFNN. In a decoder block, self-attention layer and encoder-decoder attention layer depicts relationship between current sequence and previous sequence, and relationship between current sequence and embedding. In encoder-decoder attention layer
\begin{equation} 
    Q =W_{Q} X_D,K =W_{K} X_E,V =W_{V} X_E.
\end{equation}
Here, $X_D$ denotes the input is the output of last decoder block, and $X_E$ denotes the input is the output of encoder.

Transformer models sequential property via position encoding. Suppose $i$ denotes the position of sub-sequence, and $\lambda$ denotes the dimension of sub-sequence, then the position encoding
\begin{subequations}
    \begin{gather}
        \mathrm{PE}(i,2\lambda)=\sin\left(\frac{i}{10000^{2\lambda/d}}\right),\\
        \mathrm{PE}(i,2\lambda+1)=\cos\left(\frac{i}{10000^{2\lambda/d}}\right).
    \end{gather}
\end{subequations}

Transformer is still with multi-head mechanism. When the $k$-th embedding is being decoded, only $k-1$-th and previous decoding can be seen. This multi-head mechanism is masked multi-head attention. After decoding, the output is obtained through a softmax fully-connected layer.

\subsection{Incorporating Transformer and LSTM to Kalman Filter}

\subsubsection{Motivation}

To optimize the decoded sequence, beam search \citep{beamsearch2016} was adopted in seq2seq model. Both beam search and Viterbi algorithm in hidden Markov model (HMM) are based on dynamic programming. Solving optimal estimation of current state according to observation and previous state, is called decoding or inference in HMM \citep{Robert2012MLAPP}. Beam search is an extension of Viterbi algorithm when the Markov property of state, along with conditional independence of observation, are not satisfied. 

On the other hand, Kalman Filter for linear dynamic system (LDS) is correspond to HMM \citep{emkf1996,Robert2012MLAPP}. Both of them are based on the assumptions of Markov property of state, along with conditional independence of observation. The only difference is that HMM depicts $p(x_k|x_{k-1})$ and $p(y_k|x_k)$ through transition matrix and observation matrix, while LDS usually assumpts $x_k,y_k$ follow Gaussian distribution.

Kalman Filter and Kalman Smoother are solving $p(x_k|y_{1:k})$ and $p(x_k|y_{1:N})$ respectively, which is equivalent to forward-backward algorithm in HMM. However, in LDS, the optimal state estimation (i.e. MAP estimation) can be obtained through the distribution of $x_k$ directly, while HMM adopts dynamic programming (i.e. Viterbi algorithm) for optimal estimation. Besides, both LDS and HMM estimate parameters through EM algorithm, and it is called Baum-Welch algorithm in HMM.

\subsubsection{Framework}

\begin{figure}
    \centering
    \includegraphics[width=0.6\textwidth]{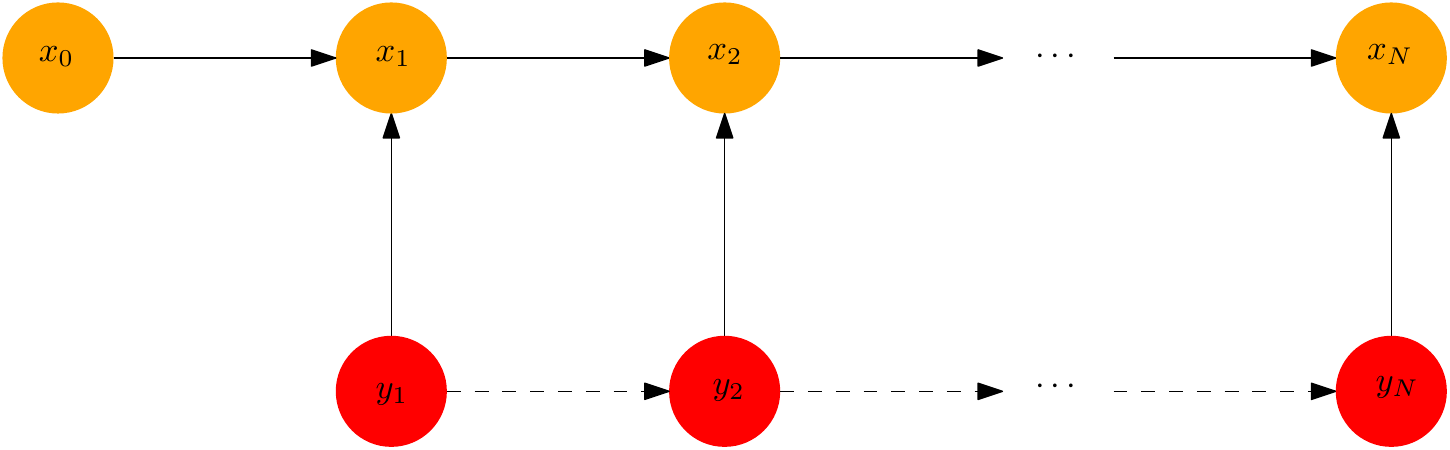}
    \caption{Basic KF solves $x_k$ via $y_k$ and $x_{k-1}$.}
    \label{fig:basickf}
\end{figure}

\begin{figure}
    \centering
    \includegraphics[width=0.85\textwidth]{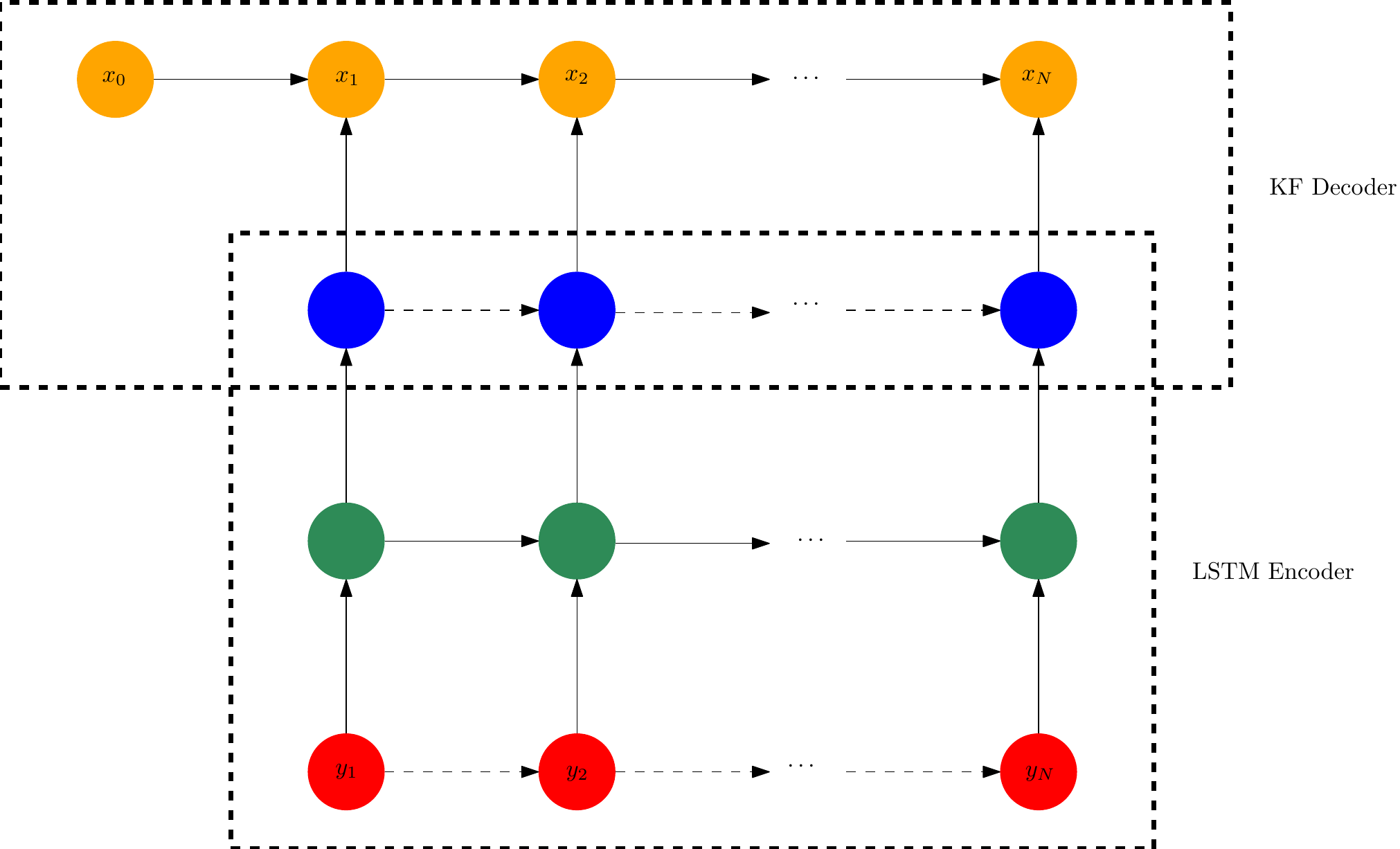}
    \caption{LSTM-KF, with LSTM encoder and KF decoder, where dashed lines denote latent connections.}\label{fig:lstmkf}
\end{figure}

Therefore, we suggest that the decoder in seq2seq model can be replaced by Kalman Fiiter. In Fig. \ref{fig:basickf}, Kalman Filter requires input and previous output for current output, which is similar to seq2seq model. Seq2seq suppress the effect of noise through encoder-decoder architecture, which is consistent to the goal of state estimation. Based on deep learning, hidden information of state is depicted more effectively, while the model would not satisfy the assumptions of Markov property of state, along with conditional independence of observation. However, as Fig. \ref{fig:lstmkf} shows, Kalman Filter receives the context from LSTM encoder, i.e. observation after LSTM processing. This model is called LSTM-KF.

Since EM-KF depends on observation only, it may not estimate parameters w.r.t. state $Q,m_0,P_0$ precisely, it is only competent to estimate $R$. Observation can not depict information of state. Deep learning for observation is capable of mining more information w.r.t. state, to enhance the performance. Hence, we proposed LSTM-KF. Based on LSTM-KF, we combined LSTM, Transformer and EM-KF, to propose Transformer-KF and TL-KF.

\begin{itemize}
    \item {\textbf{LSTM-KF:}} Input observation into LSTM, output a new series that depict state more effectively. Despite the difference of $R$ between new series and old one, as EM-KF can estimate $R$, we only need to replace old series by this new one to estimate $Q,m_0,P_0$ precisely.
    \item {\textbf{Transformer-KF:}} Transformer can capture long-term dependency (that LSTM may not) \citep{Hollis2018}, which enhance the robustness for observation with noise. Besides, Transformer is an encoder-decoder model, which is easy to incorporate into our proposed encoder-decoder framework.
    \item {\textbf{TL-KF:}} It is not sufficient to depict time-series by position encoding in Transformer. Instead, we can adopt a Transformer-LSTM structure. Attention is usually before memory in human cognitive system. The reason why Transformer can capture long-term dependency, is that it integrates multi-head self attention and residual \citep{He2016resnet} connection, and position encoding in Transformer need to be improved \citep{position2018}. Combining LSTM and Transformer can enhance both structural advantages and ability for time-series modeling of Transformer. Transformer draws on designs of convolutional neural networks (CNN). (Eg. multi-head attention vs. multi convolutional kernel, residual connection etc.) Therefore, Transformer can capture saliency that RNN may not, while RNN can better depict time-series.
\end{itemize}

\begin{figure}
    \centering
    \includegraphics[width=0.6\textwidth]{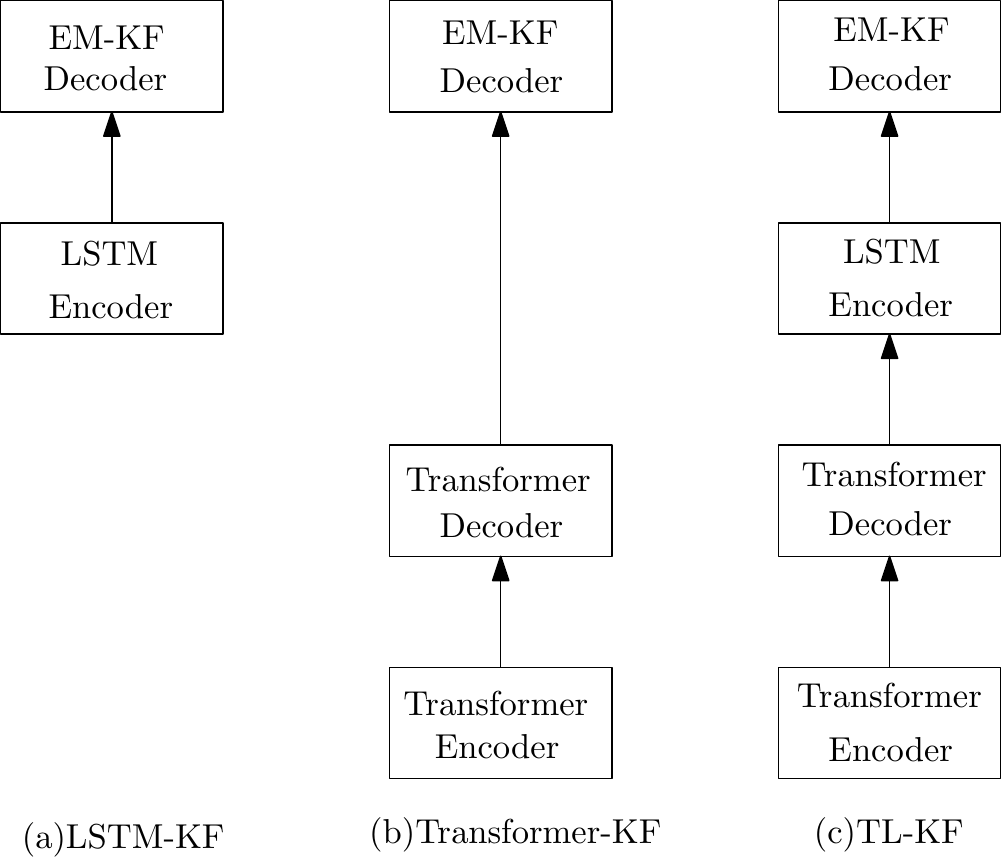}
    \caption{Deep learning models for state estimation}\label{fig:structure}
\end{figure}

The structures of LSTM-KF, Transformer-KF, and TL-KF, are shown on Fig. \ref{fig:structure}.

Table \ref{tab:method} compares encoder, decoder (inference) and parameter estimator of different models.

\begin{table}[]
    \centering
    \caption{Comparison of different models}
    \label{tab:method}
    \begin{tabular}{|c|p{3cm}|p{3cm}|p{3cm}|}
    \hline
                   & Encoder                    & Decoder (inference)                      & Parameter estimator           \\ \hline
    HMM            &                       & forward-backward algorithm, dynamic programming (Viterbi algorithm)         & EM algorithm (Baum-Welch algorithm) \\ \hline
    LSTM seq2seq   & LSTM                  & LSTM, beam search            &                \\ \hline
    KF             &                       & EM-KF                       & EM algorithm           \\ \hline
    LSTM-KF        & LSTM                  & EM-KF                       & EM algorithm           \\ \hline
    Transformer-KF & MHA + FFNN + Residual & MHA + FFNN + Residual, EM-KF & EM algorithm           \\ \hline
    \end{tabular}
\end{table}

\section{Experiments}

Consider a linear mobile robot model on one DOF, with displacement, velocity and accerelation as state variables.
\begin{equation}
    \begin{cases}
        x_k=x_{k-1}+v_{k-1}T+\frac{1}{2}a_{k-1}T^2\\
        v_k=v_{k-1}+a_{k-1}T
    \end{cases},k=1,2,...,N.
\end{equation}
Here, $T$ denotes a sample term following Shannon's sampling theorem. Suppose only displacement can be observed, thus,
\begin{equation}
    \begin{split}
        \left(\begin{array}{c}
            x_{k}\\
            v_{k}\\
            a_{k}
            \end{array}\right)&=\left(\begin{array}{ccc}
            1 & T & 0.5T^{2}\\
            0 & 1 & T\\
            0 & 0 & 1
            \end{array}\right)\left(\begin{array}{c}
            x_{k-1}\\
            v_{k-1}\\
            a_{k-1}
            \end{array}\right)+w_{k},\\
            y_{k}&=\left(\begin{array}{ccc}
            1 & 0 & 0\end{array}\right)x_{k}+v_{k},\\
            w_{k}\sim N(0,Q)&,v_{k}\sim N(0,R),x_{0}\sim N(m_{0},P_{0}). 
    \end{split} \label{eq:pnp}
\end{equation}
As $Q,R,m_{0},P_{0}$ were unknown, we first adopted EM algorithm for estimating $R$, then new series for estimating $Q,m_0,P_0$ were generated, to compare EM-KF, LSTM-KF, Transformer-KF and TL-KF. Suppose $Q,R,P_0$ are as form $\sigma^2 I$, and $m_0$is as form $(0,0,m_a)^T$. The $\sigma^2$ estimation is the quotient of the trace of covariance matrix and order of matrix, and $m_{03}$ is the estimation of $m_a$.

The iteration number of EM algorithm is 10. The de facto values $Q=1\times 10^{-2}I_3, R=5\times 10^{-3}I_1, m_0=(0,0,0.1)^T, P_0=0.1I_3$, sampling term $T=0.01$ second, sequence length $N=200$. At the beginning, we set $Q=2\times 10^{-2}I_3, R=I_1, m_0=(0,0,1)^T, P_0=5I_3$. The experiments were implemented on Python 3.7.1, with Anaconda 5.3.1 and Pytorch 1.0 (64-bit, no CUDA).

In LSTM, we adopted look back trick \citep{KIM2019} for time-series forecasting, and the look back number is 5, i.e. $o_t$ can be obtained through $o_{t-1},\cdots,o_{t-5}$. The layer number of LSTM is 3, and the size is 10. The train set and test set was divided at 9:1. The epoch number is 100.

Basic Transformer is designed for natural language processing. To adopt Transformer for processing time-series data directly, there are some settings that need to be operated. The hidden dimension of self-attention layer is the length of sequence $N$. The hidden dimension of FFNN layer is $4N$. The dimension of input $\lambda$ is the dimension of observation. The dimension of model $d=512$. Transformer is trained by introducing dropout \citep{2014Dropout}, and the dropout rate is 0.1. The head number is 4, and the epoch number is 500.

Both LSTM and Transformer were trained through Adam optimizer \citep{2014Adam}, and learning rate is 0.1.

Table \ref{tab1} gives the results of parameter estimation using these four methods. First of all, EM-KF can give a more accurate estimate of $R$. Secondly, Transformer KF gives more accurate estimates of $Q$ and LSTM-KF gives more accurate estimates of $m_0$ and $P_0$. Finally, TL-KF gives more accurate estimates of $Q, M_0$ and $P_0$. The table \ref{tab3} shows that for different initial values, the TL-KF algorithm always gives more accurate parameter estimates than the basic EM-KF algorithm.

\begin{figure}
    \centering
    \subfigure[path]{\includegraphics[width=0.45\textwidth]{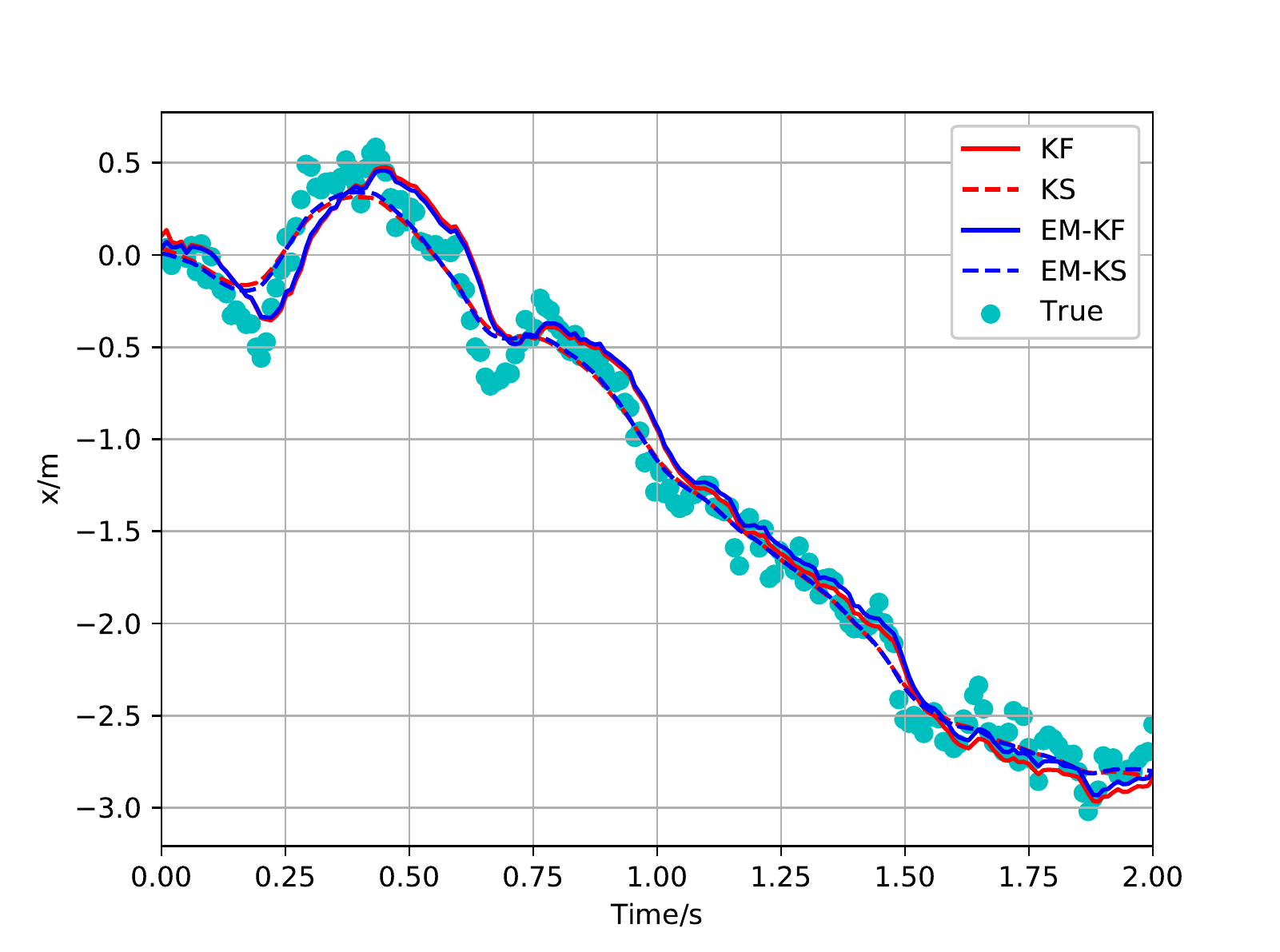}}
    \subfigure[error]{\includegraphics[width=0.45\textwidth]{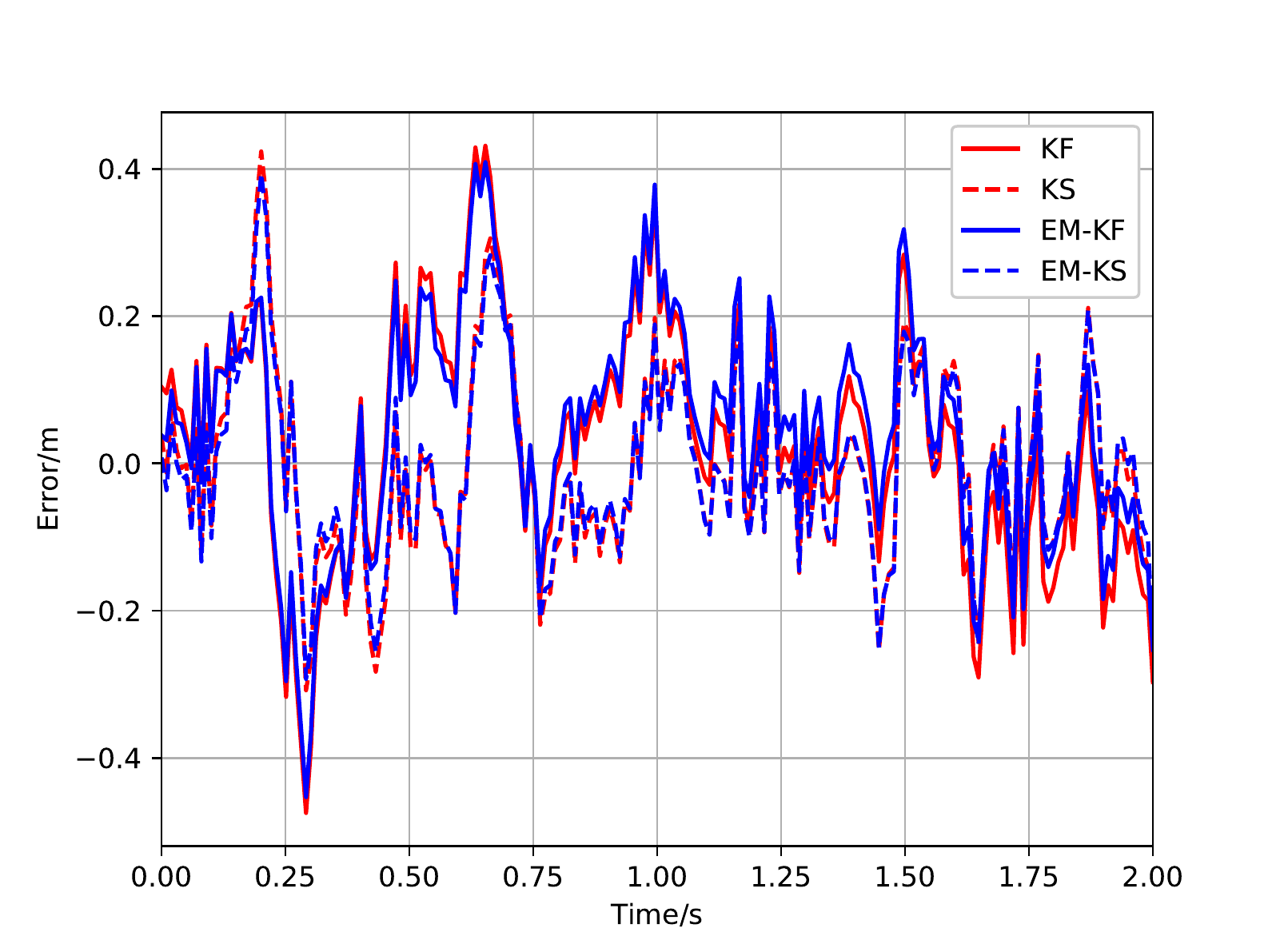}}
    \caption{EM-KF}\label{fig:em}
\end{figure}

\begin{figure}
    \centering
    \subfigure[path]{\includegraphics[width=0.45\textwidth]{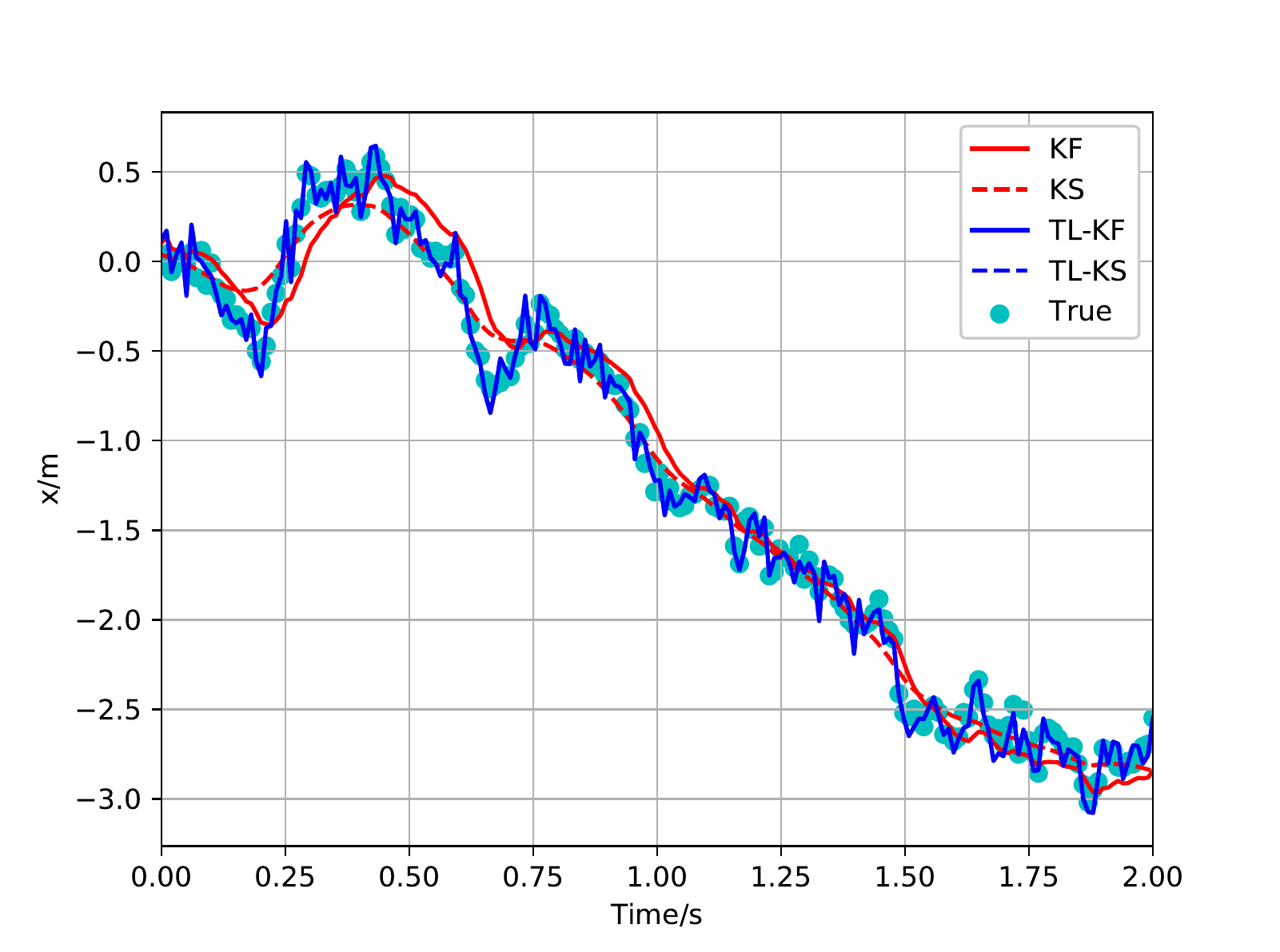}}
    \subfigure[error]{\includegraphics[width=0.45\textwidth]{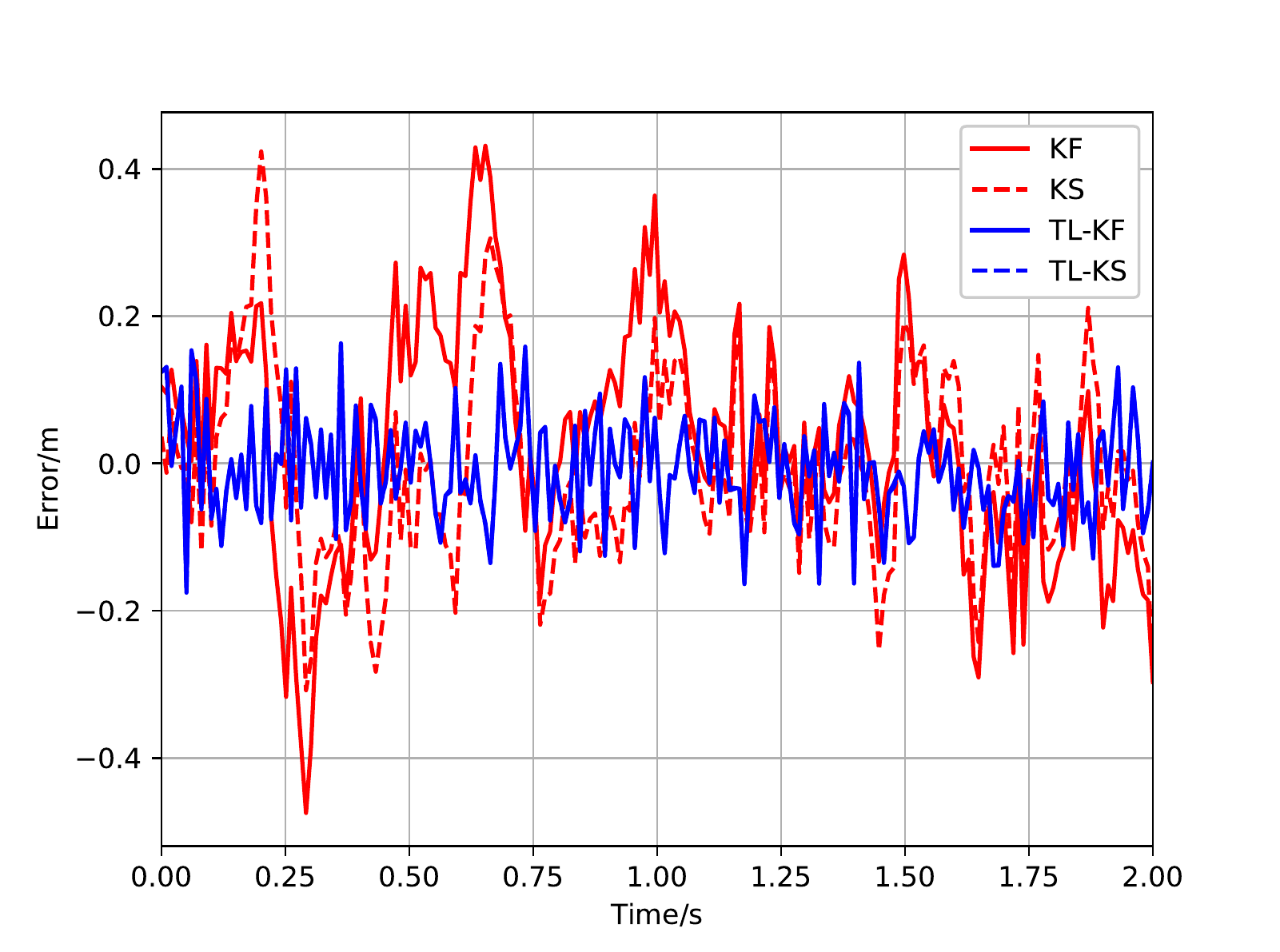}}
    \caption{TL-KF}\label{fig:tl}
\end{figure}

Table \ref{tab2} gives a comparison of the speed and accuracy between the above four methods and the basic Kalman Filter (that is, filtering directly with the initial guess of the parameter values). The displacement-time curve and the error-time curve of the above four methods are shown in Fig. \ref{fig:em}$\sim$\ref{fig:tl}, where KF and KS represent Kalman Filter and Kalman Smoother respectively, and are distinguished by solid and dotted lines in the figures.

It can be seen that for the system shown in \eqref{eq:pnp} and the initial values of system parameter iteration given in this paper, the accuracy of using EM-KF algorithm is not significantly different from that of using Kalman Filter directly. After using LSTM or Transformer to process the observation sequence, the errors can be significantly reduced. And TL-KF is superior to using LSTM or Transformer alone. Transformer alone is superior to LSTM alone in terms of speed and accuracy. The accuracy of Kalman Smoother is higher than that of Kalman Filter, but for TL-KF, the accuracy difference between filtering estimation and smoothing estimation is very small.

\begin{table}[]
    \centering
    \caption{Comparison on parameters estimation. (1) EM-KF estimates $R$ accurately. (2) Transformer-KF estimates $Q$ accurately. (3) LSTM-KF estimates $m_0,P_0$ accurately. (4) TL-KF estimates $Q,m_0,P_0$ accurately.}
    \label{tab1}
    \begin{tabular}{@{}ccccc@{}}
    \toprule
                & $\sigma_q^2$         & $\sigma_r^2$         & $m_a$ & $\sigma_p^2$           \\ \midrule
    de facto values         & $1.0\times 10^{-2}$  & $5.0\times 10^{-3}$  & 0.1   & 0.1  \\
    EM-KF          & $2.04\times 10^{-2}$ & $4.23\times 10^{-3}$ & 0.879 & 0.417\\
    LSTM-KF        & $0.14\times 10^{-2}$ & $34.3\times 10^{-3}$ & 0.109 & 0.116\\
    Transformer-KF & $1.18\times 10^{-2}$ & $98.4\times 10^{-3}$ & 0.768 & 0.359\\
    TL-KF          & $1.27\times 10^{-2}$ & $44.8\times 10^{-3}$ & 0.121 & 0.118\\ \bottomrule
    \end{tabular}
\end{table}

\begin{table}[]
    \centering
    \caption{Change the initial setting of $m_0,P_0$, then estimate parameters. TL-KF is more robust than EM-KF.}
    \label{tab3}
    \begin{tabular}{@{}cccccc@{}}
    \toprule
    Method & initial $m_a$ & initial $\sigma_p^2$ & estimating $\sigma_q^2$ & estimating $m_a$ & estimating $\sigma_p^2$ \\ \midrule
    EM & 1       & 5              & 0.0204         & 0.879   & 0.417          \\
    EM & 0.5     & 1              & 0.3447     & 0.794   & 0.239          \\
    EM & 1.5     & 15             & 0.0255         & 1.195   & 0.403          \\
    TL & 1       & 5              & 0.0126         & 0.121   & 0.118          \\
    TL & 0.5     & 1              & 0.0123         & 0.091   & 0.097          \\
    TL & 1.5     & 15             & 0.0128         & 0.089   & 0.120         \\ \bottomrule
    \end{tabular}
\end{table}

\begin{table}[]
    \centering
    \caption{Comparison on state estimation, where MSE denotes mean square error.}
    \label{tab2}
    \scalebox{0.85}{
    \begin{tabular}{@{}ccccc@{}}
    \toprule
                & Training time (sec) & EM time (sec) & Filter MSE($m^2$)          & Smoother MSE($m^2$)          \\ \midrule
    KF          &    &    & $26.83\times 10^{-3}$ & $16.95\times 10^{-3}$ \\
    EM-KF       &    & 3.98    & $26.32\times 10^{-3}$ & $15.19\times 10^{-3}$ \\
    LSTM-KF  & 76.82   & 4.15    & $17.51\times 10^{-3}$ & $11.52\times 10^{-3}$ \\
    Transformer-KF & 26.71   & 4.22    & $9.05\times 10^{-3}$  & $6.55\times 10^{-3}$  \\
    TL-KF & 91.46   & 4.23    & $5.07\times 10^{-3}$  & $4.89\times 10^{-3}$  \\ \bottomrule
    \end{tabular}}
\end{table}

\section{Conclusions and Future Works}

Combining Transformer and LSTM as an encoder-decoder framework for observation, can depict state more effectively, attenuate noise interference, and weaken the assumption of Markov property of states, and conditional independence of observations. Experiments demonstrate that our proposed model can enhance the preciseness and robustness of state estimation. Transformer, based on multi-head self attention and residual connection, can capture long-term dependency, while LSTM-encoder can model time-series. TL-KF, a combination of Transformer, LSTM and EM-KF, is precise for state estimation in systems with unknown parameters. Kalman Smoother is superior to Kalman Filter, but in TL-KF, filtering is precise enough. Therefore, after offline training for estimating $Q,m_0,P_0$, Kalman Filter for online estimation can be adopted.

The method proposed in this paper can be further studied in the following directions:
\begin{enumerate}
\item Since there are corresponding EM algorithms for EKF, UKF and particle filtering, the method in this paper can be extended to the nonlinear case, but this realization is not given in this paper, which is the main shortcoming. In addition, on the basis of combining LSTM and Kalman filtering, \citet{voden2019} proposed a unified framework for the identification and state estimation of nonlinear systems, namely the variational ordinary differential equation network (Voden).
\item For the EM-KF algorithm with initial value dependence problem, the literature \citep{Robert2012MLAPP} lists several improved methods, including the use of the subspace method \citep{Overschee1996Subspace}, variational Bayesian method \citep{Barber2006variationalbayes} and Gibbs Sampling \citep{Fruhwirthschnatter2006gibbs}. 
\item LSTM is good at mining series-related features. Combining LSTM with Transformer enhances its ability to mine specific point-related features. Similarly, LSTM can be linked to CNN \citep{cnn1989} or GAN (Generative Adversarial Network) \citep{gan2014}, which are good at capturing the salient features of a particular point \citep{lstmcnn2017,lstmgan2018}.
\item Bidirectional Encoder Representation Transformer (BERT) \citep{Devlin2018BERT} also performs better than the basic Transformer. There is a method \citep{bertpretrain2019} to significantly reduce the training time of Transformer (BERT). The bidirectional mechanism is similar to Kalman smoothing, and it is worth considering whether the bidirectional mechanism can be introduced to further improve the filtering performance.
\end{enumerate}

\begin{figure}
    \centering
    \subfigure[path]{\includegraphics[width=0.45\textwidth]{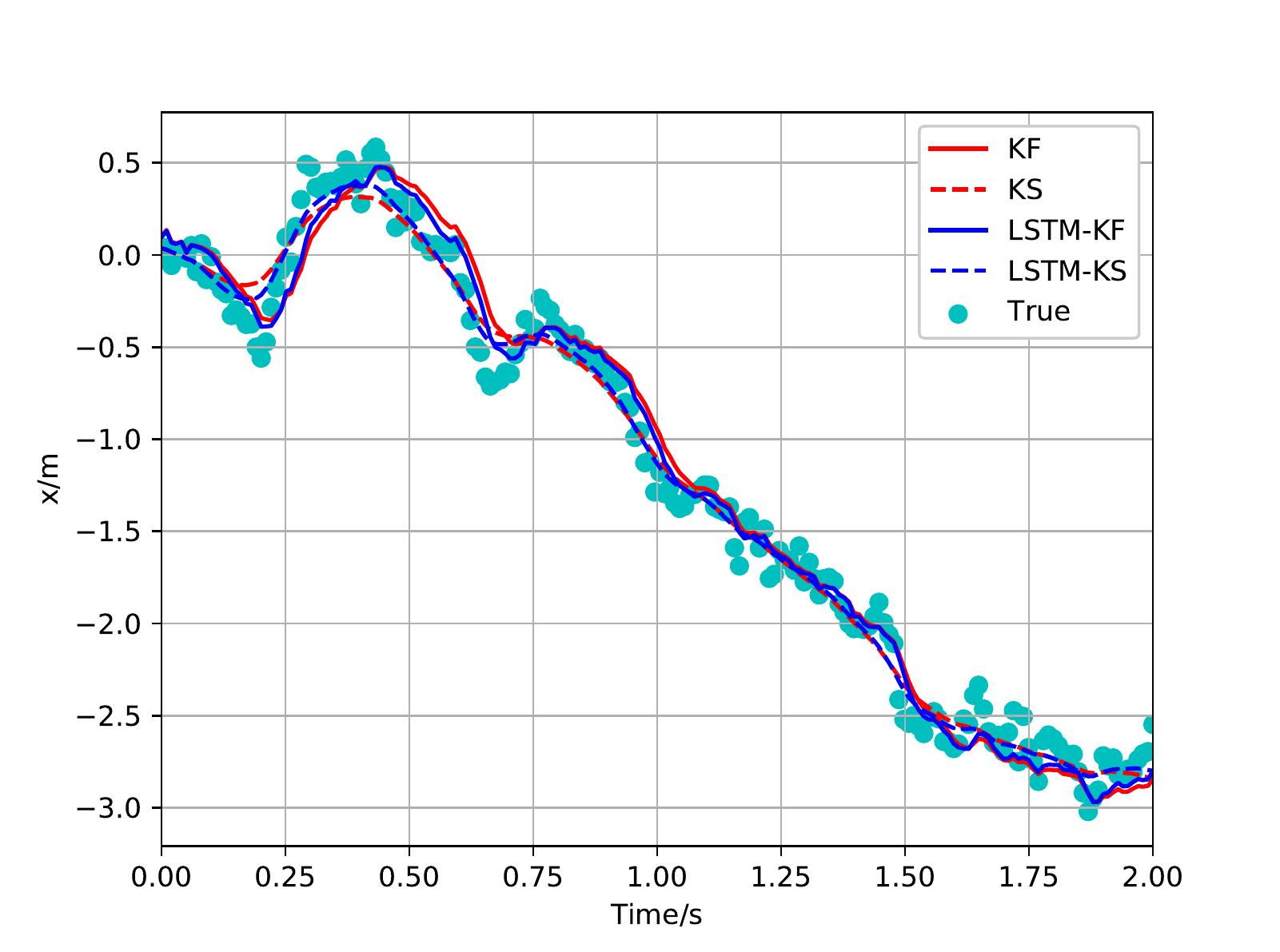}}
    \subfigure[error]{\includegraphics[width=0.45\textwidth]{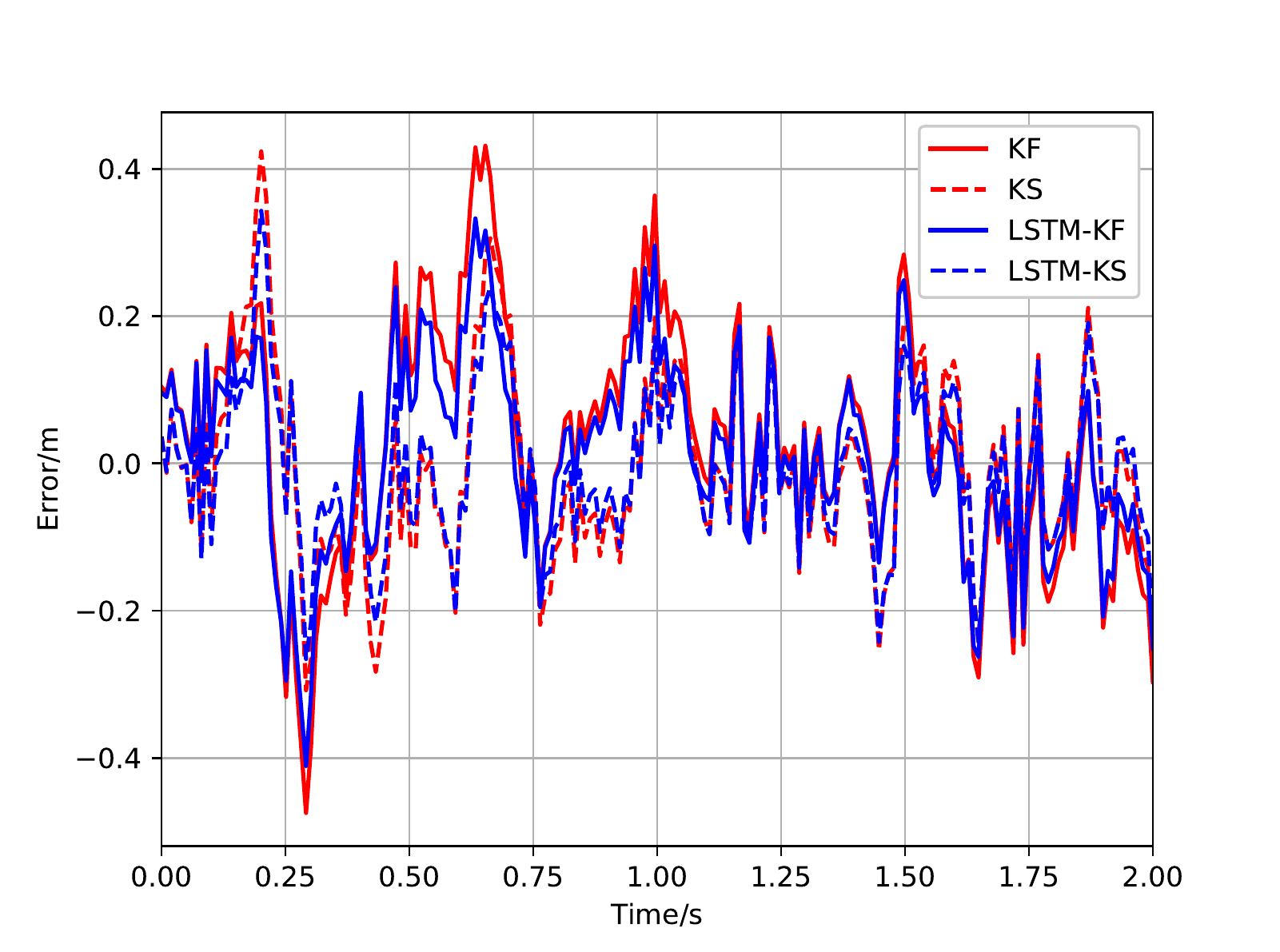}}
    \caption{LSTM-KF}\label{fig:lstm}
\end{figure}

\begin{figure}
    \centering
    \subfigure[path]{\includegraphics[width=0.45\textwidth]{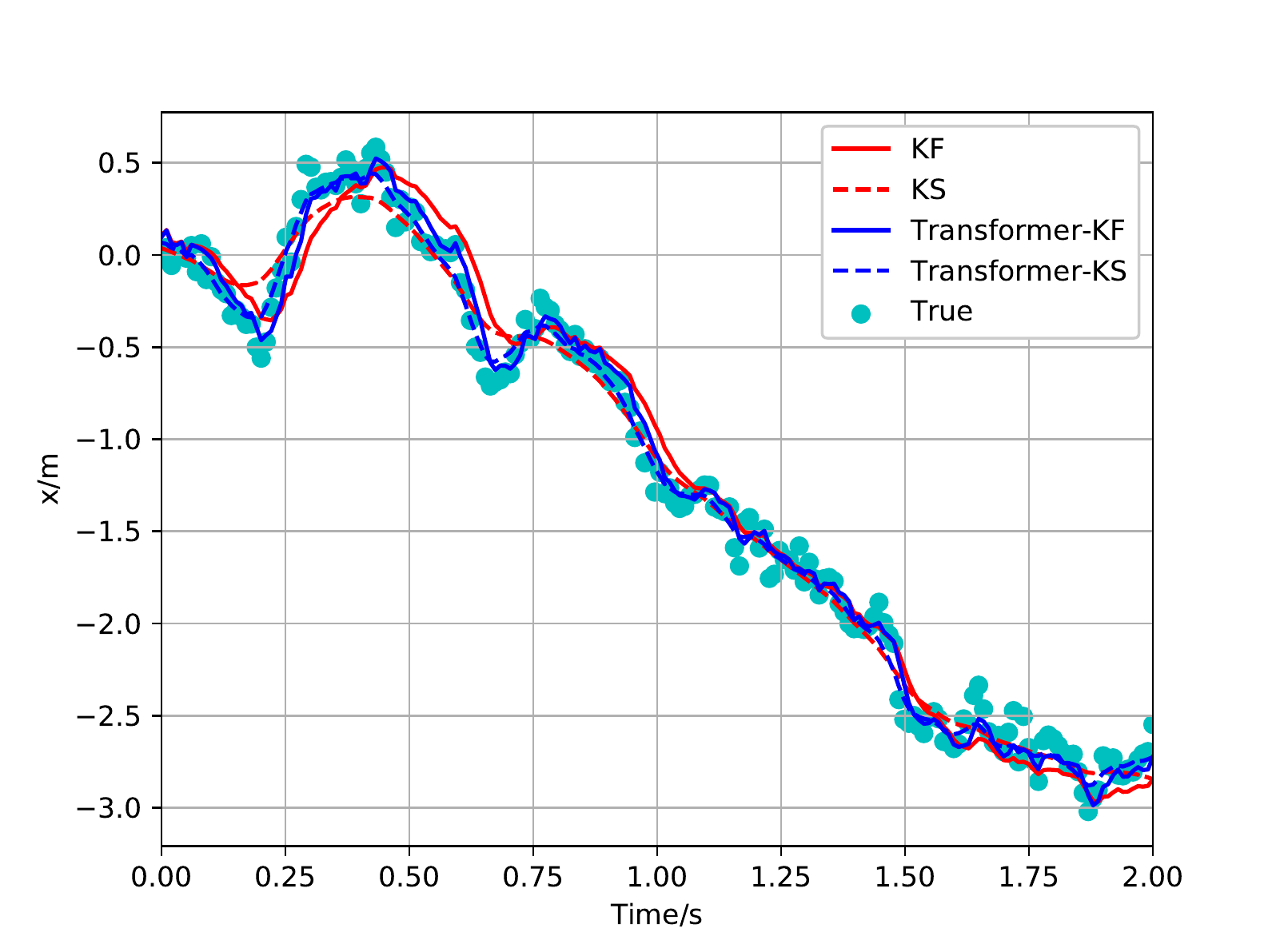}}
    \subfigure[error]{\includegraphics[width=0.45\textwidth]{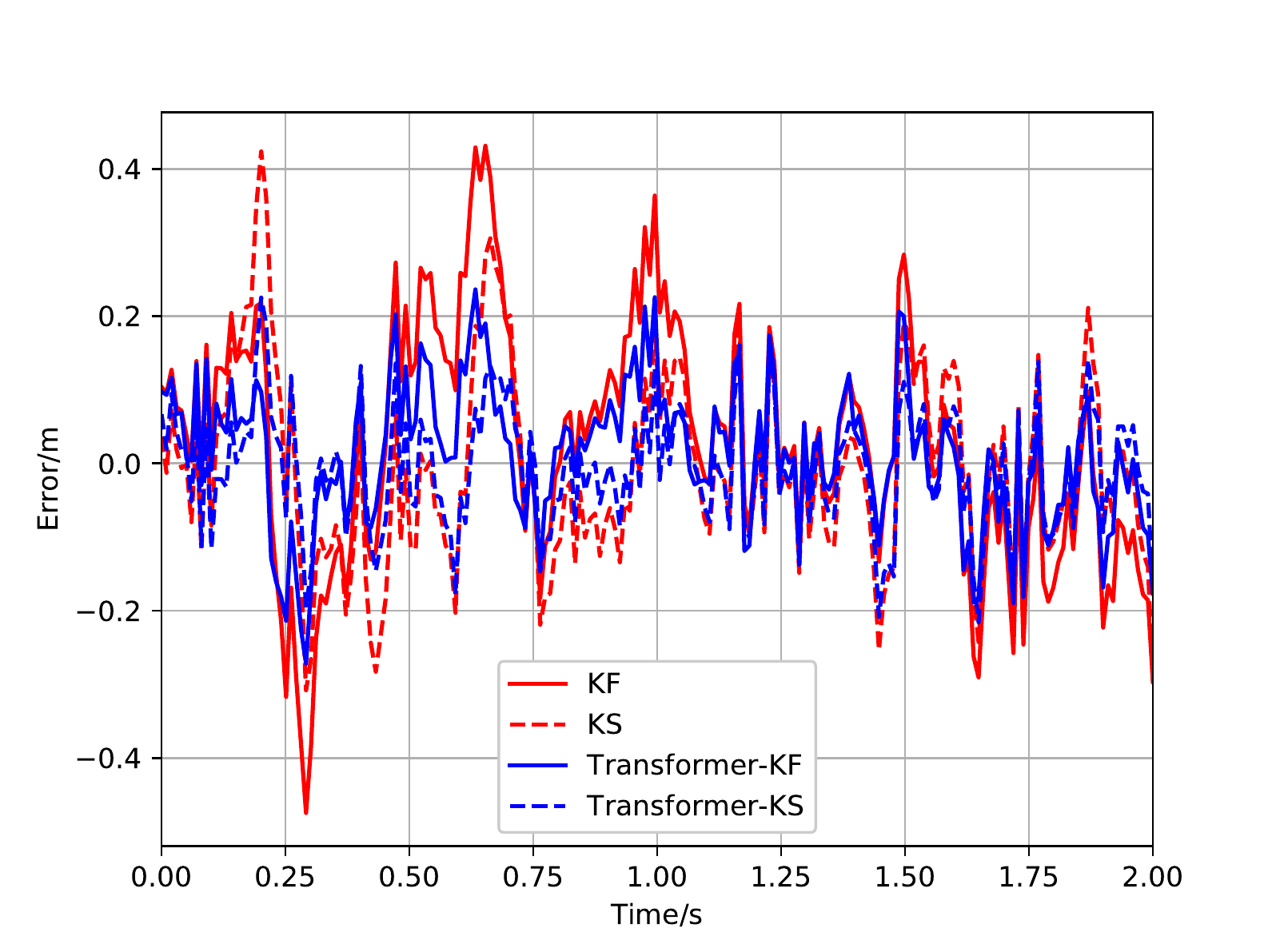}}
    \caption{Transformer-KF}\label{fig:tranf}
\end{figure}

\bibliographystyle{plainnat}
\bibliography{nkthesis}

\end{document}